\pdfoutput=1
\documentclass[11pt]{article}

\usepackage{ACL2023}
\usepackage{times}
\usepackage{latexsym}

\usepackage[T1]{fontenc}
\usepackage[utf8]{inputenc}

\usepackage{microtype}
\usepackage{inconsolata}

\usepackage{algorithm}%
\usepackage[noend]{algpseudocode}%
\usepackage{tikz}%
\usepackage{pgfplots}%
\usepackage{array}
\usepackage{pgfplotstable}%
\pgfplotsset{compat=newest}%
\usepgfplotslibrary{fillbetween}%
\usepackage{multirow}%
 \usepackage{booktabs}%
 \usepackage{graphicx}%
 \usepackage{lscape}%
\usepackage{longtable}%
\usepackage{amsmath,amsfonts,bm}

\def\eqref#1{equation~\ref{#1}}

\def\1{\bm{1}}

\def\eps{{\epsilon}}

\def\va{{\bm{a}}}

\def\vs{{\bm{s}}}

\def\evc{{c}}

\DeclareMathAlphabet{\mathsfit}{\encodingdefault}{\sfdefault}{m}{sl}
\SetMathAlphabet{\mathsfit}{bold}{\encodingdefault}{\sfdefault}{bx}{n}

\def\sC{{\mathbb{C}}}

\def\sL{{\mathbb{L}}}

\def\sS{{\mathbb{S}}}

\def\sW{{\mathbb{W}}}

\newcommand{\R}{\mathbb{R}}

\DeclareMathOperator*{\argmax}{arg\,max}
\DeclareMathOperator*{\argmin}{arg\,min}

\newcommand\blfootnote[1]{%
  \begingroup
  \renewcommand\thefootnote{}\footnote{#1}%
  \addtocounter{footnote}{-1}%
  \endgroup
}
\title{DARE: Towards Robust Text Explanations in Biomedical\\and Healthcare Applications}%
 \author{%
 	Adam Ivankay\\%
 	IBM Research Zurich\\%
 	R\"{u}schlikon, Switzerland\\%
 	\texttt{aiv@zurich.ibm.com}%
 	\And
 	Mattia Rigotti\\%
 	IBM Research Zurich\\%
 	R\"{u}schlikon, Switzerland\\%
 	\texttt{mrg@zurich.ibm.com}%
	\AND%
	Pascal Frossard\\%
	\'{E}cole Polytechnique F\'{e}d\'{e}rale de Lausanne (EPFL)\\%
	Lausanne, Switzerland\\%
	\texttt{pascal.frossard@epfl.ch}%
}%

\begin{document}%
\maketitle%
%
\begin{abstract}%
Along with the successful deployment of deep neural networks in several application domains, the need to unravel the black-box nature of these networks has seen a significant increase recently. Several methods have been introduced to provide insight into the inference process of deep neural networks. However, most of these explainability methods have been shown to be brittle in the face of adversarial perturbations of their inputs in the image and generic textual domain. In this work we show that this phenomenon extends to specific and important high stakes domains like biomedical datasets. In particular, we observe that the robustness of explanations should be characterized in terms of the accuracy of the explanation in linking a model's inputs and its decisions -- \textit{faithfulness} -- and its relevance from the perspective of domain experts -- \textit{plausibility}. This is crucial to prevent explanations that are inaccurate but still look convincing in the context of the domain at hand. To this end, we show how to adapt current attribution robustness estimation methods to a given domain, so as to take into account domain-specific plausibility. This results in our \textsc{DomainAdaptiveAREstimator} (DARE) attribution robustness estimator, allowing us to properly characterize the domain-specific robustness of faithful explanations. Next, we provide two methods, adversarial training and FAR training, to mitigate the brittleness characterized by DARE, allowing us to train networks that display robust attributions. Finally, we empirically validate our methods with extensive experiments on three established biomedical benchmarks.%
\end{abstract}%

\section{Introduction}%
\label{sec:intro}%
Research in explainable AI (XAI) has seen a surge in recent years. XAI methods aim to provide insight into the inference process and the causal links between inputs and outputs of deep neural networks (DNNs). This is pivotal in addressing many aspects of DNNs, such as fairness, potential biases and scopes of safe deployment. Especially in safety-critical domains, such as healthcare, faithful and robust explanations \citep{faithfulness} accompanying the predictions of DNNs are key to enable their deployment and understand potential false predictions and risks. Beside faithfulness, which quantifies the accuracy by which the explanations characterize the true decision-making process of the model, a second property of explanations that has been highlighted as important is plausibility \citep{socialfaithfulness, concepttransformers, faithfulnessplausibility}. Plausibility quantifies the ``degree to which some explanation is aligned with the user's understanding'' \citep{socialfaithfulness}. As such, plausibility tells us whether an explanation is found convincing and informative for domain experts. These independent but related properties are therefore crucial ingredients to provide explanations that are accurate and robust, as well as domain-relevant and convincing \citep{faithfulnessplausibility}.\par%
\blfootnote{
Code for DARE can be found at:
\url{https://github.com/ibm/domain-adaptive-attribution-robustness}
}%
Attribution methods such as Saliency Maps \citep{saliencymap}, DeepLIFT \citep{deeplift} or Integrated Gradients \citep{integratedgradients} highlight the input features that are deemed important in the decision process as heat maps. These are especially useful, as such maps are easy to interpret and no specific domain knowledge is needed to provide them. Moreover, methods like Integrated Gradients fulfill several desiderata of faithful explanations \citep{faithfulness}, which makes them an even more attractive option to explain DNNs.\par%
However, recent work has shown that attribution maps do not necessarily fulfill the \textit{robustness} aspect of faithful explanations. In particular, it has been shown in the vision domain that small input alterations can be crafted so as to change the attribution maps drastically, while leaving the prediction unchanged \citep{interpretationfragile}. Very recently, the same phenomenon has been confirmed in the textual domain as well \citep{ivankay2022fooling}.\par%
But what about the plausibility of the explanations? This paper starts by first pointing out that the importance of plausibility has been overlooked in favor of exclusively focusing on faithfulness, in particular in the textual domain. This is important because, when trying to protect a system from adversarial attacks against explanations, it is not only crucial to quantify their faithfulness, but also the plausibility of the possible adversarial samples. In fact, unfaithful but plausible explanation attacks --- \textit{convincing lies} --- have been pointed out to be particularly pernicious, since they are more difficult for domain experts to spot than equally unfaithful and implausible perturbations and explanations --- \textit{unconvincing lies} --- \citep{faithfulnessplausibility}. These observations are crucial for the use of AI explainability in high stakes scenarios, as in automated medical diagnosis, EHR classification or triage \citep{aida}, a medical professional might overlook some critical areas in a cancer cell image or disregard certain important words because they do not appear relevant according to an otherwise \textit{plausible explanation}.\par%
In this work, we focus on the robustness aspect of faithful and plausible attributions in biomedical text classification problems. Specifically, we investigate how to extract in-context adversarial perturbations which are plausible in each specific task domain under consideration. Then, we propose our attribution robustness (AR) estimator that quantifies AR in a domain-specific way. Finally, we explore methods to mitigate the domain-specific fragility of explanation methods in order to train text classifiers that can safely be deployed in safety-critical use cases like healthcare. We summarize our contributions as follows:%
\begin{itemize}%
    \item We conceptually relate faithfulness and plausibility to domain-specific attribution robustness estimation on textual data.%
    \item To this end, we extend previous work on AR estimation and introduce our AR estimator, \textsc{DomainAdaptiveAREstimator} (DARE), based on domain-plausible attacks that can be used to estimate AR in a domain-specific way.%
    \item We then empirically show that attribution maps are susceptible to adversarial perturbations that are plausible in the biomedical domain on three (multilabel) medical datasets.
    \item We are the first to develop and empirically validate two methods to mitigate adversarial perturbations and train text classifiers with robust attribution methods.%
\end{itemize}%
\begin{table*}[ht]
\centering
    \begin{tabular}{c p{0.28\textwidth} p{0.28\textwidth} p{0.28\textwidth}}
        &\centering\arraybackslash \small \textbf{\textsc{Vanilla}} & \centering\arraybackslash \small \textbf{\textsc{Adversarial}} & \centering\arraybackslash \small \textbf{\textsc{FAR-IG}} \\
        \hline\hline
        
        \rotatebox[origin=c]{90}{\textit{Original}}&%
        \begin{tabular}[c]{@{}p{0.28\textwidth}@{}}%
        \scriptsize\raggedright%
        \textcolor[rgb]{0,0,0}{'took }\textcolor[rgb]{0,0,0}{zoloft }\textcolor[rgb]{0,0,0}{for }\textcolor[rgb]{0,0,0}{5 }\textcolor[rgb]{0,0,0}{months. }\textcolor[rgb]{0.84,0.38,0.30}{\textbf{no }}\textcolor[rgb]{0.26,0.58,0.76}{\textbf{side }}\textcolor[rgb]{0,0,0}{effects }\textcolor[rgb]{0,0,0}{except }\textcolor[rgb]{0.84,0.38,0.30}{\textbf{sexual }}\textcolor[rgb]{0,0,0}{dysfunction. }\textcolor[rgb]{0.26,0.58,0.76}{\textbf{i }}\textcolor[rgb]{0,0,0}{didn't }\textcolor[rgb]{0,0,0}{feel }\textcolor[rgb]{0,0,0}{much }\textcolor[rgb]{0.84,0.38,0.30}{\textbf{better }}\textcolor[rgb]{0.13,0.40,0.67}{\textbf{or }}\textcolor[rgb]{0.84,0.38,0.30}{\textbf{{\underline{happier}} }}\textcolor[rgb]{0.84,0.38,0.30}{\textbf{and }}\textcolor[rgb]{0.26,0.58,0.76}{\textbf{it }}\textcolor[rgb]{0,0,0}{made }\textcolor[rgb]{0,0,0}{me }\textcolor[rgb]{0,0,0}{feel }\textcolor[rgb]{0,0,0}{really }\textcolor[rgb]{0,0,0}{drowsy.' }%
        \\[0.0cm]%
        \centering\arraybackslash \small $F$($\vs, \, l=$``4.0'') = 1.0%
        \end{tabular}%
        &%
        \begin{tabular}[c]{@{}p{0.28\textwidth}@{}}%
        \scriptsize\raggedright%
        \textcolor[rgb]{0,0,0}{'took }\textcolor[rgb]{0,0,0}{zoloft }\textcolor[rgb]{0,0,0}{for }\textcolor[rgb]{0,0,0}{5 }\textcolor[rgb]{0,0,0}{months. }\textcolor[rgb]{0.84,0.38,0.30}{\textbf{no }}\textcolor[rgb]{0.26,0.58,0.76}{\textbf{side }}\textcolor[rgb]{0,0,0}{effects }\textcolor[rgb]{0,0,0}{except }\textcolor[rgb]{0.84,0.38,0.30}{\textbf{sexual }}\textcolor[rgb]{0,0,0}{dysfunction. }\textcolor[rgb]{0.26,0.58,0.76}{\textbf{i }}\textcolor[rgb]{0,0,0}{didn't }\textcolor[rgb]{0,0,0}{feel }\textcolor[rgb]{0,0,0}{much }\textcolor[rgb]{0.84,0.38,0.30}{\textbf{better }}\textcolor[rgb]{0.13,0.40,0.67}{\textbf{or }}\textcolor[rgb]{0.84,0.38,0.30}{\textbf{{\underline{happier}} }}\textcolor[rgb]{0.84,0.38,0.30}{\textbf{and }}\textcolor[rgb]{0.26,0.58,0.76}{\textbf{it }}\textcolor[rgb]{0,0,0}{made }\textcolor[rgb]{0,0,0}{me }\textcolor[rgb]{0,0,0}{feel }\textcolor[rgb]{0,0,0}{really }\textcolor[rgb]{0,0,0}{drowsy.' }%
        \\[0.0cm]%
        \centering\arraybackslash \small $F$($\vs, \, l=$``4.0'') = 1.0%
        \end{tabular}%
        &%
        \begin{tabular}[c]{@{}p{0.28\textwidth}@{}}%
        \scriptsize\raggedright%
        \textcolor[rgb]{0,0,0}{'took }\textcolor[rgb]{0,0,0}{zoloft }\textcolor[rgb]{0,0,0}{for }\textcolor[rgb]{0,0,0}{5 }\textcolor[rgb]{0,0,0}{months. }\textcolor[rgb]{0.84,0.38,0.30}{\textbf{no }}\textcolor[rgb]{0.26,0.58,0.76}{\textbf{side }}\textcolor[rgb]{0,0,0}{effects }\textcolor[rgb]{0,0,0}{except }\textcolor[rgb]{0.84,0.38,0.30}{\textbf{{\underline{sexual}} }}\textcolor[rgb]{0,0,0}{dysfunction. }\textcolor[rgb]{0.26,0.58,0.76}{\textbf{i }}\textcolor[rgb]{0,0,0}{didn't }\textcolor[rgb]{0,0,0}{feel }\textcolor[rgb]{0,0,0}{much }\textcolor[rgb]{0.84,0.38,0.30}{\textbf{better }}\textcolor[rgb]{0.13,0.40,0.67}{\textbf{or }}\textcolor[rgb]{0.84,0.38,0.30}{\textbf{happier }}\textcolor[rgb]{0.84,0.38,0.30}{\textbf{and }}\textcolor[rgb]{0.26,0.58,0.76}{\textbf{it }}\textcolor[rgb]{0,0,0}{made }\textcolor[rgb]{0,0,0}{me }\textcolor[rgb]{0,0,0}{feel }\textcolor[rgb]{0,0,0}{really }\textcolor[rgb]{0,0,0}{drowsy.' }%
        \\[0.0cm]%
        \centering\arraybackslash \small $F$($\vs, \, l=$``4.0'') = 1.0%
        \end{tabular}%
        \\%
        \hline%
        
        \rotatebox[origin=c]{90}{\textit{Adversarial}}&%
        \begin{tabular}[c]{@{}p{0.28\textwidth}@{}}%
        \scriptsize\raggedright%
        \textcolor[rgb]{0,0,0}{'took }\textcolor[rgb]{0,0,0}{zoloft }\textcolor[rgb]{0,0,0}{for }\textcolor[rgb]{0.26,0.58,0.76}{\textbf{5 }}\textcolor[rgb]{0.26,0.58,0.76}{\textbf{months. }}\textcolor[rgb]{0,0,0}{no }\textcolor[rgb]{0.26,0.58,0.76}{\textbf{side }}\textcolor[rgb]{0,0,0}{effects }\textcolor[rgb]{0,0,0}{except }\textcolor[rgb]{0.26,0.58,0.76}{\textbf{sexual }}\textcolor[rgb]{0.7,0.09,0.17}{\textbf{dysfunction. }}\textcolor[rgb]{0.84,0.38,0.30}{\textbf{i }}\textcolor[rgb]{0,0,0}{didn't }\textcolor[rgb]{0.84,0.38,0.30}{\textbf{feel }}\textcolor[rgb]{0.84,0.38,0.30}{\textbf{much }}\textcolor[rgb]{0,0,0}{better }\textcolor[rgb]{0.7,0.09,0.17}{\textbf{or }}\textcolor[rgb]{0.13,0.40,0.67}{\textbf{{\underline{anything}} }}\textcolor[rgb]{0,0,0}{and }\textcolor[rgb]{0.26,0.58,0.76}{\textbf{it }}\textcolor[rgb]{0,0,0}{made }\textcolor[rgb]{0.84,0.38,0.30}{\textbf{me }}\textcolor[rgb]{0,0,0}{feel }\textcolor[rgb]{0.84,0.38,0.30}{\textbf{really }}\textcolor[rgb]{0,0,0}{drowsy.' }%
        \\[0.0cm]%
        \centering\arraybackslash \small $F$($\vs_{\mathrm{adv}}, \, l=$``4.0'') = 1.0%
        \\%
        \centering\arraybackslash \small \textbf{Cos.} = -0.32%
        \\%
        \centering\arraybackslash \small $MedSTS$ = 0.99%
        \end{tabular}%
        &%
        \begin{tabular}[c]{@{}p{0.28\textwidth}@{}}%
        \scriptsize\raggedright%
        \textcolor[rgb]{0,0,0}{'took }\textcolor[rgb]{0,0,0}{zoloft }\textcolor[rgb]{0.84,0.38,0.30}{\textbf{for }}\textcolor[rgb]{0.26,0.58,0.76}{\textbf{5 }}\textcolor[rgb]{0,0,0}{months. }\textcolor[rgb]{0.84,0.38,0.30}{\textbf{no }}\textcolor[rgb]{0,0,0}{side }\textcolor[rgb]{0,0,0}{effects }\textcolor[rgb]{0,0,0}{except }\textcolor[rgb]{0.26,0.58,0.76}{\textbf{sexual }}\textcolor[rgb]{0.7,0.09,0.17}{\textbf{dysfunction. }}\textcolor[rgb]{0.84,0.38,0.30}{\textbf{i }}\textcolor[rgb]{0.84,0.38,0.30}{\textbf{didn't }}\textcolor[rgb]{0.13,0.40,0.67}{\textbf{feel }}\textcolor[rgb]{0,0,0}{much }\textcolor[rgb]{0,0,0}{better }\textcolor[rgb]{0,0,0}{or }\textcolor[rgb]{0.26,0.58,0.76}{\textbf{{\underline{stronger}} }}\textcolor[rgb]{0.84,0.38,0.30}{\textbf{and }}\textcolor[rgb]{0.26,0.58,0.76}{\textbf{it }}\textcolor[rgb]{0,0,0}{made }\textcolor[rgb]{0,0,0}{me }\textcolor[rgb]{0,0,0}{feel }\textcolor[rgb]{0,0,0}{really }\textcolor[rgb]{0,0,0}{drowsy.' }%
        \\[0.0cm]%
        \centering\arraybackslash \small $F$($\vs_{\mathrm{adv}}, \, l=$``4.0'') = 1.0%
        \\%
        \centering\arraybackslash \small \textbf{Cos.} = -0.05%
        \\%
        \centering\arraybackslash \small $MedSTS$ = 0.96%
        \end{tabular}%
        &%
        \begin{tabular}[c]{@{}p{0.28\textwidth}@{}}%
        \scriptsize\raggedright%
        \textcolor[rgb]{0,0,0}{'took }\textcolor[rgb]{0,0,0}{zoloft }\textcolor[rgb]{0,0,0}{for }\textcolor[rgb]{0,0,0}{5 }\textcolor[rgb]{0,0,0}{months. }\textcolor[rgb]{0.84,0.38,0.30}{\textbf{no }}\textcolor[rgb]{0.26,0.58,0.76}{\textbf{side }}\textcolor[rgb]{0.84,0.38,0.30}{\textbf{effects }}\textcolor[rgb]{0,0,0}{except }\textcolor[rgb]{0.26,0.58,0.76}{\textbf{{\underline{nerve}} }}\textcolor[rgb]{0.13,0.40,0.67}{\textbf{dysfunction. }}\textcolor[rgb]{0.13,0.40,0.67}{\textbf{i }}\textcolor[rgb]{0.13,0.40,0.67}{\textbf{didn't }}\textcolor[rgb]{0.84,0.38,0.30}{\textbf{feel }}\textcolor[rgb]{0,0,0}{much }\textcolor[rgb]{0.84,0.38,0.30}{\textbf{better }}\textcolor[rgb]{0.13,0.40,0.67}{\textbf{or }}\textcolor[rgb]{0.84,0.38,0.30}{\textbf{happier }}\textcolor[rgb]{0.84,0.38,0.30}{\textbf{and }}\textcolor[rgb]{0,0,0}{it }\textcolor[rgb]{0,0,0}{made }\textcolor[rgb]{0,0,0}{me }\textcolor[rgb]{0,0,0}{feel }\textcolor[rgb]{0,0,0}{really }\textcolor[rgb]{0,0,0}{drowsy.' }%
        \\[0.0cm]%
        \centering\arraybackslash \small $F$($\vs_{\mathrm{adv}}, \, l=$``4.0'') = 1.0%
        \\%
        \centering\arraybackslash \small \textbf{Cos.} = 0.79%
        \\%
        \centering\arraybackslash \small $MedSTS$ = 0.93
        \end{tabular}%
        \\%
        \hline%
        \hline%
    \end{tabular}
    \caption{Attribution methods in medical text classifiers on the Drug Reviews dataset \citep{drugreviews}, trained without any robust objectives (\textsc{Vanilla}) are susceptible to imperceptible word substitutions. By changing one words in the original sample (underlined), the words with originally positive attributions (red) are assigned negative values (blue), and vice versa, while keeping the prediction confidence $F$ in the correct class unchanged. This is indicated by the \textit{Cosine Distance} (Cos.) between the explanations of original and adversarial samples. Attacks on attributions in networks trained with robust training objectives (\textsc{Adversarial} and our novel \textsc{FAR-IG}) are less successful (higher Cos. values) while also being more perceptible (lower medical semantic similarity - MedSTS - values between original and adversarial samples).}
    \label{fig:fragilityexample}
\end{table*}%
%

\section{Related work}%
\label{sec:relwork}%
Recently, attribution methods like Saliency Maps \citep{saliencymap}, Integrated Gradients \citep{integratedgradients}, DeepLIFT \citep{deeplift} or Shapley Values \citep{shapley} have been widely deployed in the medical domain for deep learning \citep{medicalxai}. These methods aim to provide insight into the inference process in DNNs. They highlight features in the input that are deemed relevant in the decision process, without requiring any domain-specific knowledge or heavy computation resources. Thus, they have been widely adopted in areas where predictions need to be accompanied by explanations, such as analysis of medical images \citep{medicalcam, medicallrp} or important symptoms that contribute to or against a given diagnosis \citep{lime, medicalxai}. The authors \citet{medicalxai} provide an extensive survey on such methods applied in several medical problem setups.\par%
The work of \citet{interpretationfragile} explore the robustness of such attribution methods and find that they are susceptible to adversarial perturbations, both in the image domain \citep{geometryblame, far}, and the text domain \citep{ivankay2022fooling, devilswork, misspellingrobustness, cea}. However, these works operate on general, non-domain-specific text. None investigate domain-specific text, such as healthcare, where most datasets possess unique vocabularies and semantics. We aim to provide insight into how current methods can be adapted to such specific technical domains.\par%
In order to mitigate the highlighted fragility of attributions in DNNs, several methods have been developed. The authors \citet{igsum, geometryblame, alignment} propose methods that smoothen the decision boundary of the classifiers, making gradients  smoother as well. The work of \citet{far} provides a general framework to perform adversarial training of attributions, successfully making attributions more robust to input perturbations. However, all of these methods have been developed for the continuous image domain. The transition of such methods to the discrete input space like text has not been investigated, nor has any novel method for text been introduced. In this work, we demonstrate how these shortcomings can be mitigated.%

\section{Background and Motivation}%
\label{sec:background}%
In this chapter, we introduce the background and motivation of our AR estimation. We define a text classifier $F$ as a function that maps a text sample $\vs$ to a label $l$ from a given set of labels $\sL$. In discrete input spaces like text, $F$ is a function composition of a non-differentiable embedding $E$ that maps the discrete inputs into a continuous domain $\R^{h \times p}$, and a differentiable classifier $f$ mapping the embeddings to the output logits $\R^{|\sL|}$. We denote $\sS = \{\, (\vs, l) \,| \vs = (w_i)_{i \in \{0...|\vs|-1 \} }, \; w_i \in \sW, \; l \in \sL, \; |\sS| = N\}$ as the set of $N$ text samples $\vs$ with a label $l \in \sL$, each containing a sequence of words $w_i$ drawn from the vocabulary $\sW$, $h$ the embedding dimension and $p$ the maximum sequence length. \textit{Attributions} are functions $\va = A(\vs, F, l)$ that assign a value to each word $w_i$ in a text sample $\vs$, indicating its importance in the DNN inference process. We sum up the attribution values of each $w_i$-s embedding, resulting in a single value for each word.\par%
\textit{Attribution robustness} is defined as the Lipschitz attribution robustness constant \cite{cea}, given in the following equation:%
\begin{equation}%
\label{eqn:ar}%
    r(\vs) {}={} \max_{\tilde{\vs} \in \mathcal{N}(\vs)} \frac{d \big[ A(\tilde{\vs}, F, l), \; A(\vs, F, l) \big]}{d_s(\tilde{\vs},\vs)}
\end{equation}%
with the prediction constraint
\begin{equation}%
\label{eqn:pred_constraint}%
    \argmax_{i \in \{1...|\sL|\}}F_i(\tilde{\vs}) {}={} \argmax_{i \in \{1...|\sL|\}}F_i(\vs)%
\end{equation}%
Here, $r(\vs)$ denotes the robustness of attribution method $A$ computed for text sample $\vs$ with label $l$, drawn from $\sL$, and classifier $F$. The function $d[A(\tilde{\vs}, F, l), A(\vs, F, l)]$ denotes the distance between original and adversarial attribution maps $A(\tilde{\vs}, F, l)$ and $A(\vs, F, l)$, $\mathcal{N}(\vs)$ is a predefined neighborhood of text sample $\vs$. The term $d_s(\tilde{\vs},\vs)$ indicates the distance of adversarial and original input texts. The robustness of an attribution method on a test dataset $\sS$ then becomes the average of $r(\vs)$ over the dataset. Note that the robustness of an attribution method on a classifier is inversely proportional to the constant computed in Equation (\ref{eqn:ar}), as large attribution distances and small input distances result in large constants, indicating low robustness. This reflects the definition of the robustness property of \textit{faithful} explanations \citep{faithfulness}.\par%
Our first contribution is conceptual and is motivated by the observation that \textit{plausibility} is a criterion that is rooted in the specific domain and the semantic conventions within it. Thus, methods to guard against adversarial attacks on explanations need to be \textit{domain adaptive} to conform to the threat model that prioritizes unfaithful explanations which are \textit{semantically plausible} in the domain under consideration, thus are particularly misleading and potentially dangerous.\par%
The strategy we propose to control \textit{domain-adaptive plausibility} is based on the observation that while the numerator in Equation (\ref{eqn:ar}) characterizes \textit{faithfulness} by quantifying the effect of adversarial attacks on attributions, the denominator can be adapted to capture \textit{plausibility} by promoting adversarial attacks that remain close to the original input in a semantically meaningful way, in the domain under consideration. In particular, while \citet{ivankay2022fooling} utilize the cosine distance of sentence embeddings obtained from domain-agnostic encoders like Universal Sentence Encoder \citep{use} and MiniLM \citep{tse}, we can obtain a domain-specific measure of distance by using embeddings trained on the domain of interest. This will control the plausibility of the adversarial samples by making sure that their domain-dependent semantic distance remains close to the original inputs.\par%
Table \ref{fig:fragilityexample} exemplifies this approach of quantifying the fragility of attributions in medical text by simultaneously keeping track of faithfulness and plausibility through domain-adapted semantic similarity.%
%
%
%

\section{Medical Attribution Robustness}%
\label{sec:medicalar}%
Current AR estimation algorithms \citep{ivankay2022fooling, devilswork} were designed to operate in the general text domain, such as news articles \cite{agnews_mr, fakenews}, movie reviews \citep{imdb} or product reviews \citep{yelp} and make use of the generously available labeled data in these domains. This section describes our proposed methods to adapt these algorithms to the biomedical and healthcare domains where data is sparse and the vocabularies are domain-specific. We describe our datasets and models, we observe that current estimators can be made domain-adaptive by abstracting the candidate extractor and finally, inspired by the works of \citet{cea}, we introduce our estimation algorithm \textsc{DomainAdaptiveAREstimator} (DARE), which can effectively be used to estimate AR in the domain of biomedical text.%
\subsection{Medical datasets}%
\label{subsec:datasetsmodels}%
In healthcare, text can appear in many different forms with diverse vocabularies. Thus, we choose three text datasets that cover different aspects of relevant use cases in the medical domain. Often, the datasets are not large enough to train models with state-of-the-art numbers of parameters, such as transformers. Therefore, we make heavy use of transfer learning by utilizing pretrained transformer-based language models and finetune them on our datasets.\par%
Our first dataset, Drug Reviews (DR) \citep{drugreviews}, consists of patient reviews of different medical drugs, classified into a rating of 1 to 10 for patient satisfaction. The dataset contains 215063 samples, written in mostly layman's terms along with the names of the drugs and symptom descriptions. Given the dataset's nature, the classification model we choose is a finetuned RoBERTa model, with pretrained weights from Hugging Face \citep{huggingface}.\par%
The Hallmarks of Cancer \citep{hoc} dataset (HoC) consists of 1852 biomedical publication abstract associated with 0 or more hallmarks of cancer \citep{hallmarksofcancer}. The samples are peer-reviewed publication texts, containing few to no misspellings with scientific biomedical vocabulary. As the dataset contains only a small amount of samples, we finetune a pretrained BioLinkBERT \citep{biolinkbert} model from Hugging Face to achieve state-of-the-art classification accuracy on this dataset.\par%
Lastly, we evaluate the MIMIC-III \citep{mimic} Discharge Summary dataset (MIMIC). This is a set of extremely long, de-identified, free text ICU discharge summaries from patients admitted to critical care, written by medical professionals. The corresponding ICD-9 codes \citep{icd9} are associated with each sample in a multilabel fashion. This dataset contains in average 2500 words per sample \citep{mimic}, thus traditional BERT-based models are not feasible as their runtime scales quadratically with the sequence length. Therefore, we finetune a pretrained Clinical-Longformer model \citep{clinicallongformer}, a Longformer MLM \citep{longformer} trained on the MIMIC-III discharge summaries. For an in-depth, more detailed description of our datasets and models, we refer to Appendix \ref{apn:modelsdatasets}.\par%
\subsection{AR in multilabel datasets}%
\label{subsec:armultilabel}
Many text classification datasets in healthcare do not only have one label per sample. In HoC, multiple hallmarks can be associated with an abstract, and  MIMIC contains hardly any discharge summary with only one associated ICD-9 code. However, current AR estimation definitions only focus on the single label case. Therefore, we make the following modifications to make AR work in the multilabel case. First, we modify the prediction constraint from Equation (\ref{eqn:ar}) to reflect multilabel predictions. The label $l$ becomes a set of predicted labels, and the prediction constraint in Equation (\ref{eqn:pred_constraint}) holds as long as the predicted set of labels from the original sample is equal to the one from the adversarial sample. We denote this constraint as P in our estimation algorithm. Second, attribution methods compute maps on a per-class basis, where the overall attribution $\mathbf{A} = A(\vs, F, l)$ equals the attribution of the single predicted class $l$. In a multilabel case, we extend this notion to the sum of attributions for each predicted class, thus the overall attribution map becomes $\displaystyle \mathbf{A} = \sum_{l_i \in l}A(\vs, F, l_i)$.
\begin{table*}[t]
\centering
\begin{tabular}{@{}l c c c@{}}
\textsc{\textbf{MLM}}                     & \textsc{HoC}    & \textsc{Drug Reviews} & \textsc{MIMIC-III}  \\\hline\hline
\textsc{BERT} \citep{bert}                   & 0.786                  & 0.702        & 0.677      \\
\textsc{DistilBERT} \citep{distilbert}             & 0.733                  & 0.599        & 0.580      \\
\textsc{DistilRoBERTa} \citep{distilroberta}          & 0.768                  & \textbf{0.745}        & 0.604      \\
\textsc{PubMedBERT} \citep{pubmedbert}             & \textbf{0.908}                  & 0.704        & 0.781      \\
\textsc{BioClinicalBERT} \citep{bioclinicalbert}       & 0.775                  & 0.629        & 0.847      \\
\textsc{ClinicalBigBird} \citep{clinicallongformer}       & -                      & -            & 0.372      \\
\textsc{Clinical-Longformer} \citep{clinicallongformer}    & -                      & -            & \textbf{0.867}\\\hline\hline
\end{tabular}%

\caption{Top-5 accuracies of the masked language models (MLMs) on our datasets Hallmarks of Cancer (HoC), Drug Reviews and MIMIC-III. Each word in each sample of the dataset is masked and the sample is then propagated through the MLM. If the original masked word is in the top-5 predictions of the MLM, the sample counts as positive.}
\label{tbl:mlmaccuracies}
\end{table*}%
\begin{algorithm}[t]
\caption{\texttt{DomainAdaptiveAREstimator}}\label{alg:dare}
\textbf{Input}: Input $\vs$ with label set $l$, classifier $F$, attribution $A$, distance metric $d$, prediction constraint P, language model MLM, number of candidates $|\sC|$, maximum perturbation word ratio $\rho_{max}$\\
\textbf{Output}: Adversarial sentence $\vs_{\mathrm{adv}}$
\begin{algorithmic}[1]
\State $\vs_\mathrm{adv} \gets \vs$, $d_{max} \gets 0$, $n \gets 0$
\State $I_{\vs} {}={} \nabla_{\vs} d \big[ A(\vs + \eps, F, l), \; A(\vs, F, l) \big]$
\For{$w_i \in \langle w_1, ..., w_{|\vs|} \rangle | I_{m-1} \geq I_{m} \forall{m} \in \{2, ..., |\vs|\}$}
    \If{$w_i \in \sS_{\mathtt{Stop\,words}}$}
        \State \textbf{continue}
    \EndIf
    \State $\sC_i \gets \mathrm{MLM}(w_i, \vs, |\sC|)$
    \For{$\evc_k \in \sC_i$}
        \State \begin{small}{$\tilde{\vs}_{w_{ik}} \gets$ Replace $w_i$ in $\vs_\mathrm{adv}$ with $\evc_k$}\end{small}
        
        \If{$\mathrm{P}(F_j(\tilde{\vs}_{w_{ik}}), l)$ not satisfied}
        	\State \textbf{continue}
        \EndIf
        
        \State \begin{small}{$\tilde{d} = d \big[ A(\tilde{\vs}_{w_{ik}}, F, l),A(\vs, F, l) \big]$}\end{small}
    
        \If{$\tilde{d} > d_{max}$}
        
            \State $\vs_\mathrm{adv} \gets \tilde{\vs}_{w_{ik}}$
            \State $d_{max} \gets \tilde{d}$
            \State $n \gets n + 1$
        \EndIf
    \If{$\rho {}={} \frac{n + 1}{|\vs|} > \rho_{max}$}
        \State \textbf{break}
    \EndIf
\EndFor
\EndFor
\end{algorithmic}
\end{algorithm}%
\subsection{DomainAdaptiveAREstimator (DARE)}%
\label{subsec:dare}%
Candidate extractors are essential parts of AR estimators, as they provide substitution candidates for the input words, largely contributing to the plausibility and perceptibility of the adversarial alterations. We find that candidate extractors in current work \citep{ivankay2022fooling, cea}, the counter-fitted synonym embeddings \citep{counterfitted} and the masked language model (MLM) DistilBERT \citep{distilbert}, are suboptimal in our case, due to their vocabulary only minimally overlapping with the ones from our datasets. However, following the idea of \citet{cea}, we argue that, when using the right model, MLMs are in fact effective candidate extractors for word substitutions. Not only do they take context of the words into account, but can be trained on unlabeled data in an unsupervised fashion, thus pretrained models are available for many domains and use cases. Therefore, they can easily be adapted to any domain, without the need for labeled synonym data. For this reason, as our substitution candidate extractors, we choose a pretrained MLM that maximizes the top-5 accuracy of predicting the words in dataset, when each is masked separately, averaged over the dataset. This metric is used as it represents how well the MLMs capture the context of the words, providing meaningful and in-context substitution candidates that will likely result in fluent adversarial samples. Consequently, we use the MLMs DistilRoBERTa \citep{distilroberta} for Drug Reviews, PubMedBERT \citep{pubmedbert} for HoC and Clinical-Longformer \citep{clinicallongformer} for MIMIC-III. Table \ref{tbl:mlmaccuracies} summarizes the accuracies of the MLMs that we have tested.\par%
In order to estimate the AR of our classifiers, we propose our two-step, domain-adaptive AR estimator, DARE, written in Algorithm \ref{alg:dare}. In the first step, an importance ranking of the words in the text samples is extracted in order to prioritize words that are \textit{likely} to impact attributions when substituted. In contrast to current work, we use the gradient of attribution distance as ranking, as this is computationally less heavy than substituting each word with the mask token and performing a single forward pass for each. The second step of DARE is then the extraction of in-context candidates for the highest ranked words, with the pretrained MLMs discussed above and substituting the words greedily with the candidate that maximizes $r(\vs)$ in Equation (\ref{eqn:ar}). This allows for efficiently characterizing the robustness aspect of faithfulness while making sure the substitutions are in-context, relevant and maintain the plausibility of attributions.%
%

\section{Robust Attributions}%
\label{sec:robustattributions}%
In this section, we describe our methods to mitigate fragility of attribution maps in text. Specifically, we are the first to introduce adversarial training \citep{advtraining} as a baseline \citep{devilswork} and our adapted FAR \citep{far} training as a novel method to achieve state-of-the-art attribution robustness in deep neural networks for text classification. Even though we describe and later evaluate the methods on biomedical datasets, these are general training methods that are applicable to any text classification problem.%
\subsection{Adversarial Training}%
\label{subsec:at}%
In an untargeted setup, adversarial training \citep{deepfool, advtraining} augments the training data with samples $\vs_{\mathrm{adv}}$ specifically computed as a function of $\vs$ to maximize the classification loss $l_c$, written in Equation (\ref{eqn:advtraining}).%
\begin{equation}%
\label{eqn:advtraining}%
  \vs_{\mathrm{adv}} = \argmax_{\tilde{\vs} \in \mathcal{N}(\vs)}  l_c(\tilde{\vs}, F, l)
\end{equation}%
where $\mathcal{N}$ denotes the search neighborhood of original sample $\vs$, $F$ the classifier and $l$ the true label of sample $\vs$. The classifiers then are trained following the optimization objective in Equation (\ref{eqn:advtrainingoptimizaiton}).%
\begin{equation}%
\label{eqn:advtrainingoptimizaiton}%
  \theta^* {}={} \argmin_{\theta}  \sum_{\vs \in \sS} l_c(\vs_{\mathrm{adv}}, F, l)
\end{equation}%
where $\theta^*$ denotes the optimal model parameters. It has been shown both in the image \citep{alignment,geometryblame,igsum} and the text domain \citep{devilswork} that adversarial training not only enhances prediction robustness in classifiers, but also improves attribution robustness.\par%
In order to solve the inner optimization problem in Equation (\ref{eqn:advtrainingoptimizaiton}), we choose the \textsc{A2T} \citep{a2t} attack framework, as it provides flexibility in terms of candidate extraction methods and is optimized for adversarial training runtime. By adapting \textsc{A2T} to use our the MLMs described in Section \ref{subsec:dare}, we successfully extract in-context and imperceptible adversarial samples for training.%
\subsection{FAR for Text}%
\label{subsec:fartext}%
The authors \citet{far} introduced a general framework for training robust attributions (FAR) in deep neural networks in the image domain. They achieve state-of-the-art robustness with few assumptions about the networks or attribution methods. Intuitively, FAR performs adversarial training on attributions and trains networks to minimize the maximal distance between original and adversarial attributions. Equation (\ref{eqn:far}) describes their extraction of adversarial samples for training.%
\begin{equation}%
\label{eqn:far}%
\begin{split}%
  \vs_{\mathrm{adv}} = &\argmax_{\tilde{\vs} \in \mathcal{N}(\vs)} \Bigl\{ ( 1 - \gamma) \cdot l_c(\tilde{\vs}, F, l)\\
        & + \gamma \cdot d \big[ A(\tilde{\vs}, F, l), \; A(\vs, F, l) \big] \Bigr\}
\end{split}%
\end{equation}%
with $\vs_{\mathrm{adv}}$ denoting the adversarial sample, $\mathcal{N}$ the neighborhood space of the original sample $\vs$, $l_c$ the classification loss of classifier $F$ on $\vs$ with true label $l$. $d$ denotes a distance between attribution maps $A$, $\gamma$ a constant with $0 \leq \gamma \leq 1$.\par%
Given the above extraction of adversarial samples, the authors train robust networks by solving the following optimization in Equation (\ref{eqn:faroptimization}).
\begin{equation}%
\label{eqn:faroptimization}%
\begin{split}%
    \theta^* {}={}  & \argmin_{\theta}  \sum_{\vs \in \sS} \Bigl\{ (1 - \delta) \cdot l_c(\vs_{\mathrm{adv}}, F, l) \\
                    & + \delta \cdot d \big[\vs_{\mathrm{adv}}, F, l), \; A(\vs, F, l) \big] \Bigr\}
\end{split}%
\end{equation}%
with the notation kept from the previous sections and $\delta$ denoting a constant with $0 \leq \delta \leq 1$.\par%
The algorithm was designed to work in the image domain. It requires each point in the embedding space (pixel space) to be a valid input. In our case, as text is a discrete input space, this does not hold. Thus, to make the method work for text inputs, we make the following adaptations. Instead of extracting the adversarial samples with the gradient-based IFIA algorithm described in the original paper, we utilize our Algorithm \ref{alg:dare} from Section \ref{subsec:dare} to solve the inner maximization in Equation (\ref{eqn:far}). To this end, the prediction constraint in Line 9 of DARE (Algorithm \ref{alg:dare}) can be omitted to allow for adversarial samples that maximize prediction loss. Moreover, the classification loss can be added as an additive term to the attribution loss in Line 11 to enable joint training of robust predictions and attributions. With our modifications, we successfully overcome the drawbacks of FAR while maintaining the benefits of training robust networks.%
%
%

\section{Experiments}%
\label{sec:experiments}%
\newcommand{\ds}[2]{\multirow{3}{*}[#2]{\rotatebox[origin=c]{90}{{\textbf{#1}}}}}
\newcommand{\res}[2]{\begin{tabular}[c]{@{}c@{}}#1\\[-0.15cm]\scriptsize{$\pm$#2}\end{tabular}}
\newcommand{\bres}[2]{\begin{tabular}[c]{@{}c@{}}\textbf{#1}\\[-0.15cm]\scriptsize{$\pm$#2}\end{tabular}}

\addtolength{\tabcolsep}{0pt}  

\begin{table*}[t]
\centering
\begin{tabular}{@{}cc|cccc|cccc|cccc@{}}%
\multicolumn{2}{c}{}                   & \multicolumn{4}{c}{${cos(A_{\mathrm{adv}}, A)}$}   &   \multicolumn{4}{c}{{\textit{MedSTS}}}   &   \multicolumn{4}{c}{{${r(\vs)}$}}     \\
& \multicolumn{1}{c|}{{\textsc{Model}}}                   & \textit{S}                   & \textit{DL}  & \textit{IG}  & \textit{A}       &   \textit{S}   &   \textit{DL}  &   \textit{IG}  &   \textit{A}       &   \textit{S}   &   \textit{DL}  &   \textit{IG}  &   \textit{A}   \\\hline\hline

\ds{HoC}{-0.3cm}
& \textsc{Van.}
& \res{0.67}{0.22}   & \res{-0.09}{0.22}    & \res{0.06}{0.27}& \res{0.66}{0.14}& \res{0.79}{0.12}& \res{0.79}{0.13}& \res{0.79}{0.09}& \res{0.78}{0.1}& \res{0.76}{0.11}& \res{2.6}{0.11}& \res{2.2}{0.22}& \res{0.77}{0.11}  \\
& \textsc{Adv.}
& \res{0.81}{0.09}   & \res{0.09}{0.22}    & \res{0.46}{0.23}& \res{0.74}{0.14}& \res{0.79}{0.1}& \res{0.79}{0.13}& \res{0.79}{0.09}& \res{0.78}{0.1}& \res{0.45}{0.11}& \res{2.2}{0.25}& \res{1.3}{0.16}& \res{0.59}{0.09}  \\
& \textsc{FAR-IG}
& \bres{0.84}{0.08}   & \bres{0.24}{0.2}    & \bres{0.65}{0.26}& \bres{0.86}{0.08}& \res{0.77}{0.14}& \res{0.77}{0.14}& \res{0.78}{0.11}& \res{0.77}{0.14}& \bres{0.35}{0.12}& \bres{1.6}{0.31}& \bres{0.8}{0.31}& \bres{0.3}{0.05}  \\\hline\hline

\ds{Drug Rev.}{-0.17cm}
& \textsc{Van.}
& \res{0.89}{0.12}   & \res{0.25}{0.32}    & \res{0.48}{0.35}& \res{0.72}{0.18}& \res{0.92}{0.08}& \res{0.92}{0.09}& \res{0.92}{0.09}& \res{0.91}{0.09}& \res{0.69}{0.07}& \res{4.1}{0.19}& \res{3.3}{0.22}& \res{2.1}{0.1}  \\
& \textsc{Adv.}
& \res{0.91}{0.12}   & \res{0.36}{0.3}    & \res{0.49}{0.34}& \res{0.78}{0.17}& \res{0.91}{0.09}& \res{0.9}{0.1}& \res{0.91}{0.09}& \res{0.9}{0.09}& \res{0.45}{0.06}& \res{3.7}{0.17}& \res{2.8}{0.14}& \res{1.1}{0.09}  \\
& \textsc{FAR-IG}
& \bres{0.93}{0.11}   & \bres{0.77}{0.28}    & \bres{0.86}{0.21}& \bres{0.86}{0.12}& \res{0.9}{0.09}& \res{0.9}{0.09}& \res{0.9}{0.09}& \res{0.89}{0.1}& \bres{0.35}{0.05}& \bres{1.2}{0.14}& \bres{0.8}{0.14}& \bres{0.73}{0.07}  \\\hline\hline

\ds{MIMIC-III}{0.07cm}
& \textsc{Van.}
& \res{0.35}{0.27}   & \res{0.08}{0.33}    & \res{0.0}{0.37}& \res{0.7}{0.26}& \res{0.88}{0.07}& \res{0.84}{0.07}& \res{0.82}{0.11}& \res{0.84}{0.07}& \res{3.1}{}& \res{2.9}{0.18}& \res{2.8}{0.15}& \res{0.94}{0.2}  \\
& \textsc{Adv.}
& \bres{0.44}{0.32}   & \bres{0.12}{0.26}    & \res{0.0}{0.45}& \bres{0.76}{0.21}& \res{0.85}{0.07}& \res{0.77}{0.19}& \res{0.8}{0.03}& \res{0.81}{0.13}& \bres{1.9}{0.21}& \bres{1.9}{0.47}& \bres{2.5}{0.27}& \bres{0.63}{0.12}  \\
& \textsc{FAR-IG}
& --   & --    & --& --& --& --& --& --& --& --& --& --\\\hline\hline

\end{tabular}

\caption{Attribution robustness metrics (mean and stddev.) of the vanilla (Van.), adversarially trained (Adv.) and FAR-trained (FAR-IG) models, trained on our three datasets. We perform AR estimation for the attributions S, DL, IG and A. The reported metrics are the cosine similarity between attributions of original and adversarial samples - ${cos(A_{\mathrm{adv}}, A)}$ -, the semantic similarity of the two input text samples - \textit{{MedSTS}} - as well as the estimated attribution robustness constant - ${r(\vs)}$ -. We conclude that the vanilla models perform poorly in terms of attribution robustness, while both adversarially and FAR-IG trained models are significantly more robust, yielding higher attribution similarities and lower ${r(\vs)}$ values. FAR-IG models outperform adversarially trained models, giving the most promising method to train attributionally robust networks.}%
\label{tbl:results}%
\end{table*}%

\addtolength{\tabcolsep}{3pt}  
In this section, we report our experiments and setup to estimate attribution robustness in the biomedical domain. We compare the robustness of four attribution methods on three text classifiers trained naively and with robust optimization objectives (adversarial training and FAR). Our results show that the naively trained models are heavily sensitive to imperceptible word substitution attacks, while the two robust training methods significantly increase attribution robustness, with FAR outperforming adversarial training.%
\subsection{Experimental setup}%
\label{subsec:setup}%
For each dataset described in Section \ref{subsec:datasetsmodels}, we compare the attribution robustness of a classification model trained with three different training objectives: i) a vanilla natural model trained with the cross-entropy loss; ii) a model trained with adversarial training as described in Section \ref{subsec:at} and iii) a model trained with robust FAR objectives from Section \ref{subsec:fartext}. The attribution methods evaluated are Saliency (S) \citep{saliencymap}, DeepLIFT (DL) \citep{deeplift}, Integrated Gradients (IG) \citep{integratedgradients} and the models' self-attention weights (A) \citep{attentionmechanism}. We choose these as they are popular methods to provide explanations for DNNs in healthcare \citep{medicalxai}. We use DARE from Section \ref{subsec:dare}, with the corresponding MLMs from Table \ref{tbl:mlmaccuracies} to extract adversarial samples and analyze the cosine distance of original and adversarial attributions, the semantic similarity between original and adversarial input text samples (using the MedSTS semantic embeddings) and combining these two metrics, the resulting attribution robustness constants $r(\vs)$, described in Section \ref{sec:background}. A complete set of estimation parameters is given in Table \ref{tbl:estimationparams} of the appendix.\par%
To evaluate the semantic similarity between original and perturbed inputs, current methods utilize state-of-the-art sentence embeddings on the STSBenchmark dataset \citep{stsbenchmark}. We argue that this is suboptimal, as it is not clear whether it captures perturbation perceptibility in the biomedical domain as well. Therefore, we utilize the model made public by \citet{medstsmodel} to evaluate semantic distance between texts. This model is the top performing RoBERTa model on the MedSTS dataset \citep{medsts}, a state-of-the-art dataset for semantic similarity in the biomedical domain.\par%
Our vanilla (Van.) models are trained with the standard cross-entropy classification loss, the adversarially trained models (Adv.) with the \textsc{A2T} adversarial training framework \citep{a2t}, utilizing the MLMs from Table \ref{tbl:mlmaccuracies} as candidate extractors. To train our FAR robust models (FAR-IG), we use the FAR training framework described in Section \ref{subsec:fartext}, using DARE to solve the inner maximization of Equation (\ref{eqn:faroptimization}), the cosine distance as attribution distance and Integrated Gradients (IG) as attribution method. For reproducibility, we report the full set of training parameters in Table \ref{apn:trainingparams}, \ref{apn:advtrainingparams} and \ref{apn:fartrainingparams}. The estimation is reported with a three-fold cross validation, averaging the results. The models and datasets are implemented in PyTorch \citep{pytorch} and PyTorch Lightning \citep{pytorchlightning}, the pretrained weights are taken from the Hugging Face library \citep{huggingface}, with the attributions implemented with Captum \citep{captum}. The models are finetuned on the datasets using 4 Nvidia A100 GPUs.%
\subsection{Results}%
\label{subsec:results}%
Table \ref{tbl:results} summarizes the results of our experiments. We observe that the non-robust vanilla models (Van.) perform poorly in terms of cosine distance between original and adversarial attribution maps compared to their robust counterparts (Adv. and FAR-IG). Especially the attributions DeepLIFT (DL) and Integrated Gradients (IG) are significantly altered by the attacks. This is reflected in the higher estimated robustness constants $r(\vs)$ for the vanilla models. Thus, we conclude that training networks with no robustness objective is largely suboptimal if faithful and robust explanations are needed.\par%
However, both the baseline adversarial training and our adapted FAR objectives are able to train networks with significantly more robust attributions than vanilla training. For the HoC dataset and IG attributions, adversarial training increases the cosine similarity up to 0.46, while FAR-IG training increases it by 0.65 over. A similar trend is observable for the other models, datasets and attribution methods. FAR-IG training reduces the estimated robustness constants consistently by ~40-60\%, which is a significant increase in robustness. This convinces us that FAR is a feasible method to achieve robust attributions in DNNs. \par%
We further observe that even if our FAR-IG model is not evaluated on IG, but on S, DL or A, it still outperforms vanilla and adversarially trained models both in terms of ${cos(A_{\mathrm{adv}}, A)}$ and $r(\vs)$. Therefore, we conclude that the robustness attained by FAR training with IG transfers to other attributions, further strengthening our confidence in FAR being an attractive option to train robust networks.%
%
%
%

\section{Conclusion}%
\label{sec:conclusion}%
In this work, we explored the attribution robustness of biomedical text classification. We extended current robustness estimators to introduce DARE, a domain-adaptive AR estimator. Then, we showed on three different biomedical datasets that classifiers trained without robust objectives lack robustness to small input perturbations in this domain as well. In order to mitigate this, we proposed two training methods, adversarial training and FAR to train neural networks that yield robust attributions even in the presence of carefully crafted input perturbations. With our experiments, we show that adversarial training and FAR are able to increase the attribution robustness significantly, with FAR giving the best results.\par%
Our work is a key milestone for the deployment of DNNs in the biomedical domain, as such a safety-critical application area requires sound and faithful explanations. In the future, we plan to extend our investigation from text classification to other NLP problems in the biomedical domain. Moreover, investigating the robustness of other types explanation methods is an important future research direction.%

\section{Limitations and Risks}%
DARE only works for text. In its introduced form, it requires the prediction gradients for importance ranking, thus can only be used to attack differentiable architectures (up to the embedding layer). Most state-of-the-art classifiers (DNNs, transformers) fulfill this criteria though. Moreover, DARE requires MLMs trained in a specific domain to work -- which might not always be readily available. However, as MLMs can be trained in an unsupervised fashion, pretrained MLMs can be finetuned to that domain with rather low effort.\par%
The main risk of DARE is that it does not give a guaranteed lower bound of robustness. If an attacker develops a stronger attack that is able to compute better perturbations that alter attributions to a greater extent, having a model that is robust to DARE perturbations might not be sufficient to withstand those stronger attacks. Taking the robustness estimation for granted is a risk, as it is true for most other attacks in traditional adversarial setups. This directly indicates another risk, namely that DARE could be used to attack explanations in deployed systems that are not trained robustly.\par%
We train our methods on state-of-the-art Nvidia A100 GPUs. Without having such GPUs available, FAR training in particular becomes a bottleneck, as the computation graph needs to be stored for several forward and backward passes, depending on the attribution method used. On this end, we also require the attributions to be differentiable with respect to the input embeddings, which is an implicit requirement of the FAR training method. We do not see any risks in using FAR to train robust networks.\par%
Finally, we do not examine any other aspects of faithful interpretations, only the robustness. We assume that these methods reflect the model behavior to some extent, but do not conclude any experiments to verify this assumption. Further investigation into whether more robust attributions yield better faithfulness in other aspects could be an interesting future research topic.%
%

\bibliographystyle{acl_natbib}
\bibliography{bibliography}

\onecolumn
\newpage
\appendix
\setcounter{page}{1}
\section{Appendix}
\subsection{Models and Datasets}%
\label{apn:modelsdatasets}%
We use three public datasets to evaluate the attribution robustness of biomedical text classifiers. Our main goal is to show how robust attribution methods are on these datasets, thus we do not aim to advance the state-of-the-art for classification accuracy, but train models that achieve close to state-of-the-art performance while being relatively easy to train. For each dataset, we use a 60\%-20\%-20\% split for training, test and validation splits, apply basic preprocessing by lower casing the text, removing characters that are not in the Latin alphabet and remove double spaces, new line symbols and double quotes.\par%
The Drug Reviews (DR) dataset consists of patient reviews of different medical drugs, classified into a rating of 1 to 10 for patient satisfaction. In order to increase classification performance, we reduce the number of classes to 5 by merging classes 1 and 2, 3 and 4, 5 and 6 etc. The dataset contains 215063 samples, and we train a RoBERTa model for classification, with the standard cross entropy loss on the first 128 tokens.\par%
The Hallmarks of Cancer (HoC) dataset comprises 1852 biomedical publication abstract associated with 0 or more hallmarks of cancer, thus is a 10-class multilabel classification dataset. We finetune a pretrained BioLinkBERT model for classification, use the first 256 tokens as inputs to the model after tokenization and utilize the binary cross entropy as classification weight.\par%
Our last dataset, the MIMIC-III Discharge Summary dataset consists of patients' ICU discharge summaries, associated with their ICD-9 codes. In order to reduce the overall number of classes from ~1800, we only take the 50 most frequent ICD-9 codes. This results in a total of 59647 samples. As the summaries are very long, we finetune a pretrained Clinical-Longformer model for classification, with a maximum sequence length of 4096, default attention window size and global attention on the \texttt{[CLS]} token.\par%
Table \ref{apn:modeldesc} summarizes our models, Table \ref{apn:trainingparams} contains the used hyperparameters for our finetuning process and Table \ref{apn:accuracies} the resulting accuracies of all our trained models. We use the AdamW optimizer throughout all our experiments.\par%
The Hallmarks of Cancer and Drug Reviews dataset are publicly available datasets. The requirements for MIMIC-III \footnote{https://physionet.org/content/mimiciii/1.4/} were completed and we comply with their DUA.
\begin{table*}[ht]
\small
\centering
\begin{tabular}{lccc}
\textbf{\textsc{Parameter}}                    & \textsc{Hallmarks of Cancer}              & \textsc{Drug Reviews}     & \textsc{MIMIC-III}                        \\\hline\hline
\textsc{Input shape}             & (256,)                                    & (128,)                    & (4096,)                                   \\
\textsc{Num. classes}            & 10                                        & 5                         & 50                                        \\
\textsc{HF Model ID}   & \small michiyasunaga/BioLinkBERT-base & \small roberta-base   & \small yikuan8/Clinical-Longformer              \\
\textsc{Num. params}             &  108240394 & 124649477    & 148697906\\\hline\hline 
\end{tabular}
\caption{Parameters of our classification models.}
\label{apn:modeldesc}
\end{table*}%
\begin{table*}[t]
\small
\centering
\begin{tabular}{lccc}
\textbf{\textsc{Parameter}}                                                               & \textsc{Hallmarks of Cancer}                                                        & \textsc{Drug Reviews}  & \textsc{MIMIC-III}                                                                 \\\hline\hline
\begin{tabular}[c]{@{}l@{}}\textsc{Classification}\\ \textsc{loss}\end{tabular}         & \begin{tabular}[c]{@{}c@{}}Multilabel binary\\ cross entropy\end{tabular} & Cross entropy & \begin{tabular}[c]{@{}c@{}}Multilabel binary\\ cross entropy\end{tabular} \\
\textsc{LR}                                                                    & 0.00001                                                                   & 0.000001      & 0.00004                                                                   \\
\textsc{Batch size}                                                            & 128                                                                       & 64            & 4                                                                         \\
\textsc{Epochs}                                                                & 50                                                                        & 50            & 50                                                                        \\
\textsc{Precision}                                                             & 32                                                                        & 32            & 16                                                                        \\
\begin{tabular}[c]{@{}l@{}}\textsc{Accumulate}\\ \textsc{gradient batches}\end{tabular} & 1                                                                         & 1             & 4   \\\hline\hline                                                                  
\end{tabular}
\caption{Parameters used to train our non-robust, vanilla models.}
\label{apn:trainingparams}
\end{table*}%
\begin{table*}[ht]
\small
\centering

\begin{tabular}{clccccccccc}
                                                       &           & \multicolumn{3}{c}{\textsc{Hallmarks of Cancer}} & \multicolumn{3}{c}{\textsc{Drug Reviews}} & \multicolumn{3}{c}{\textsc{MIMIC-III}} \\\hline\hline
                                                       & \textsc{\textbf{Model}}     & \textit{Van.}      & \textit{Adv.}      & \textit{FAR-IG}      & \textit{Van.}        & \textit{Adv.}        & \textit{FAR-IG}       & \textit{Van.}                        & \textit{Adv.}                        & \textit{FAR-IG}                       \\\hline
\multirow{5}{*}{\rotatebox[origin=c]{90}{{\textbf{\textsc{Nat.}}}}}     & \textsc{Accuracy}  & 0.95      & 0.94      & 0.92        & 0.9         & 0.92        & 0.92         & 0.92                        &  0.9                         & -                            \\
                                                       & \textsc{Precision} & 0.78      & 0.74      & 0.62        & 0.89        & 0.92        & 0.92         & 0.59             &        0.57                  & -                            \\
                                                       & \textsc{Recall}    & 0.89      & 0.82      & 0.90        & 0.9         & 0.92        & 0.92         & 0.71                 &     0.61                 & -                            \\
                                                       & \textsc{F1-score}  & 0.82      & 0.78      & 0.73        & 0.9         & 0.92        & 0.92         & 0.64               &    0.6                   & -                            \\
                                                       & \textsc{Loss}      & 0.24      & 0.27      & 0.27        & 0.68        & 0.36        & 0.32         & 0.3                &     0.33                    & -                            \\\hline
\multirow{5}{*}[0.03cm]{\rotatebox[origin=c]{90}{{\textbf{\textsc{Adv.}}}}} & \textsc{Accuracy}  & 0.88      & 0.89      & 0.87        & 0.61        & 0.67        & 0.65         &     0.89       &     0.9                    & -                            \\
                                                       & \textsc{Precision} & 0.55      & 0.59      & 0.5         & 0.61        & 0.66        & 0.65         &    0.54       &      0.55                  & -                            \\
                                                       & \textsc{Recall}    & 0.75      & 0.7       & 0.8         & 0.6         & 0.67        & 0.65         &    0.67          &    0.61                   & -                            \\
                                                       & \textsc{F1-score}  & 0.61      & 0.63      & 0.62        & 0.6         & 0.67        & 0.65         &   0.59        &     0.62                  & -                            \\
                                                       & \textsc{Loss}      & 0.64      & 0.53      & 0.44        & 2.5         & 1.1         & 1.2          &     0.41        &   0.39                    & -  \\\hline\hline                          
\end{tabular}
\caption{Natural (\textbf{\textsc{Nat.}}) and adversarial (\textbf{\textsc{Adv.}}) classification metrics of the non-robust (Van.), adversarially robust (Adv.) and FAR-trained (FAR-IG) models. All metrics are macro-averaged over the samples, as our datasets are highly class-imbalanced.}
\label{apn:accuracies}
\end{table*}%
\subsection{AR Estimation and Robust Training}%
\label{apn:robusttraining}%
%
%
%
\begin{table*}[ht]%
\small
\centering%
\begin{tabular}{@{}lccc@{}}%
\textsc{\textbf{Parameter}}                                                           & \textsc{Hallmarks of Cancer}    & \textsc{Drug Reviews}  & \textsc{MIMIC-III}            \\\hline\hline
\begin{tabular}[c]{@{}l@{}}\textsc{Candidate}\\ \textsc{extractor}\end{tabular}   & PubMedBERT            & DistilRoBERTa & Clinical-Longformer    \\
\textsc{$\rho_{\mathrm{max}}$}                                           & 0.05                  & 0.05          & 0.005                 \\
\textsc{$|\sC|$}                                                         & 5                     & 5             & 3                     \\
\textsc{$d(A_{\mathrm{adv}}, A)$}                                               & cosine                & cosine        & cosine                \\
\textsc{$d_s(\vs_{\mathrm{adv}}, \vs)$} & \begin{tabular}[c]{@{}c@{}}MedSTS semantic\\embeddings\end{tabular} & \begin{tabular}[c]{@{}c@{}}MedSTS semantic\\embeddings\end{tabular} & \begin{tabular}[c]{@{}c@{}}MedSTS semantic\\embeddings\end{tabular}\\\hline\hline 
\end{tabular}%
\caption{Hyperparameters used for estimating attribution robustness for our three datasets Hallmarks of Cancer, Drug Reviews and MIMIC-III. Candidate extractor denotes the MLM used for extracting the replacement candidates in DARE, $\rho_{\mathrm{max}}$ the maximum ratio of perturbed words in each sample, $|\sC|$ the number of replacement candidates extracted for each word, $d(A_{\mathrm{adv}}, A)$ the attribution distance metric and $d_s(\vs_{\mathrm{adv}}, \vs)$ the text input distance.}%
\label{tbl:estimationparams}%
\end{table*}
In order to achieve robust attributions, in addition to the vanilla models we train models with robust training objectives. During adversarial training, we augment the training batches with adversarial samples that maximize classification loss. We use the \textsc{A2T} training method for extracting adversarial samples, with the parameters summarized in Table \ref{apn:advtrainingparams}. Our FAR models are trained with the robust objectives from Section \ref{subsec:fartext}, and the hyperparameters are written in Table \ref{apn:fartrainingparams}.
\begin{table*}[ht]
\small
\centering
\begin{tabular}{lccc}
\textsc{\textbf{Parameter}}                                                                            & \textsc{Hallmarks of Cancer}    & \textsc{Drug Reviews}  & \textsc{MIMIC-III}            \\\hline\hline
\begin{tabular}[c]{@{}c@{}}\textsc{Candidate}\\ \textsc{extractor}\end{tabular}                   & PubMedBERT            & DistilRoBERTa & Clinical-Longformer    \\
$\rho_{\mathrm{max}}$                                                           & 0.05                  & 0.05          & 0.005                 \\
$|\sC|$                                                                         & 5                     & 5             & 3                     \\
\begin{tabular}[c]{@{}l@{}}\textsc{Classification}\\ \textsc{loss}\end{tabular}                   & \begin{tabular}[c]{@{}l@{}}Multilabel binary\\ cross entropy\end{tabular} & Cross entropy & \begin{tabular}[c]{@{}c@{}}Multilabel binary\\ cross entropy\end{tabular} \\

\begin{tabular}[c]{@{}l@{}}\textsc{Ratio of attacked}\\ \textsc{samples in batch}\end{tabular}    & 0.3                   & 0.3           & 0.3                   \\
\textsc{LR}                                                                              & 0.00001               & 0.000001      &   0.000001    \\
\textsc{Batch size}                                                                      & 32                    &   64          &   16  \\
\textsc{Epochs}                                                                          & 30                    &   20          &   20                                                          \\\hline\hline 
\end{tabular}
\caption{Parameters used to train our adversarially robust networks.}
\label{apn:advtrainingparams}
\end{table*}%
\begin{table}[t]
\small
\centering
\begin{tabular}{lcc}
\textbf{\textsc{Parameter}}                                                                     & \textsc{Hallmarks of Cancer}                                                        & \textsc{Drug Reviews}  \\\hline\hline
\begin{tabular}[c]{@{}l@{}}\textsc{Candidate}\\ \textsc{extractor}\end{tabular}                & PubMedBERT                                                                & DistilRoBERTa \\
$\rho_{\mathrm{max}}$                                                        & 0.05                                                                      & 0.05          \\
$A$                                                                          & IG                                                                       & IG            \\
$d(A_{\mathrm{adv}}, A)$                                                            & cosine                                                                       & cosine           \\
$|\sC|$                                                                      & 5                                                                         & 5             \\
\begin{tabular}[c]{@{}l@{}}\textsc{Classification}\\ \textsc{loss}\end{tabular}                & \begin{tabular}[c]{@{}l@{}}Multilabel binary\\ cross entropy\end{tabular} & Cross entropy \\
\begin{tabular}[c]{@{}l@{}}\textsc{FAR}\\ \textsc{instantiation}\end{tabular}                                                          & AdvAAT                                                                    & AAT           \\
$\gamma$                                                                     & 0.85                                                                       & 0.0           \\
$\delta$                                                                     & 0.85                                                                       & 0.7           \\
\textsc{LR}                                                                           & 0.00001                                                                   & 0.000001      \\
\textsc{Batch size  }                                                                 & 4                                                                         & 8             \\
\textsc{Epochs}                                                                       & 30                                                                        & 20            \\
\begin{tabular}[c]{@{}l@{}}\textsc{Ratio of attacked}\\ \textsc{samples in batch}\end{tabular} & 0.6                                                                       & 0.6          \\\hline\hline 
\end{tabular}
\caption{Parameters used to train our FAR-IG networks.}
\label{apn:fartrainingparams}
\end{table}%
\onecolumn%
\subsection{More Examples}%
\label{apn:moreexamples}%
\input{src_appendix/more_examples.tex}%


\end{document}